\documentclass[10pt,twocolumn,letterpaper]{article}

\usepackage{iccv}
\usepackage{times}
\usepackage{epsfig}
\usepackage{graphicx}
\usepackage{amsmath}
\usepackage{amssymb}
\usepackage{bbm}
\usepackage{makecell}
\usepackage{booktabs}
\usepackage{multirow}
\usepackage{multicol}

\usepackage[accsupp]{axessibility}  % Improves PDF readability for those with disabilities.

\usepackage[pagebackref=false,breaklinks=true,letterpaper=true,colorlinks,bookmarks=false]{hyperref}

\iccvfinalcopy % *** Uncomment this line for the final submission

\def\iccvPaperID{5508} % *** Enter the ICCV Paper ID here

\ificcvfinal\fi

\begin{document}

%%%%%%%%% TITLE
\title{Low-Light Image Enhancement with Illumination-Aware Gamma Correction and Complete Image Modelling Network}

\author{
    Yinglong Wang\textsuperscript{\rm1\footnotemark[1]} \qquad Zhen Liu\textsuperscript{\rm2\thanks{Equal contribution}} \qquad Jianzhuang Liu\textsuperscript{\rm3} \qquad Songcen Xu\textsuperscript{\rm4} \qquad Shuaicheng Liu\textsuperscript{\rm5,\rm2}\thanks{Corresponding author}\\
    \textsuperscript{\rm1} Meituan Inc. \ 
     \textsuperscript{\rm2}Megvii Technology \ \textsuperscript{\rm3}Shenzhen Institute of Advanced Technology \\
    \textsuperscript{\rm4}Huawei Noah’s Ark Lab \
    \textsuperscript{\rm5}University of Electronic Science and Technology of China \\
    \tt\small ylwanguestc@gmail.com, \tt\small liuzhen03@megvii.com, \tt\small jz.liu@siat.ac.cn \\ \tt\small xusongcen@huawei.com, \tt\small liushuaicheng@uestc.edu.cn
}

\maketitle
\ificcvfinal\thispagestyle{empty}\fi

%%%%%%%%% ABSTRACT
\begin{abstract}
   This paper presents a novel network structure with illumination-aware gamma correction and complete image modelling to solve the low-light image enhancement problem. Low-light environments usually lead to less informative large-scale dark areas, directly learning deep representations from low-light images is insensitive to recovering normal illumination. We propose to integrate the effectiveness of gamma correction with the strong modelling capacities of deep networks, which enables the correction factor gamma to be learned in a coarse to elaborate manner via adaptively perceiving the deviated illumination. Because exponential operation introduces high computational complexity, we propose to use Taylor Series to approximate gamma correction, accelerating the training and inference speed. Dark areas usually occupy large scales in low-light images, common local modelling structures, e.g., CNN, SwinIR, are thus insufficient to recover accurate illumination across whole low-light images. We propose a novel Transformer block to completely simulate the dependencies of all pixels across images via a local-to-global hierarchical attention mechanism, so that dark areas could be inferred by borrowing the information from far informative regions in a highly effective manner. Extensive experiments on several benchmark datasets demonstrate that our approach outperforms state-of-the-art methods.
\end{abstract}

%%%%%%%%% BODY TEXT
\section{Introduction}
\label{sec:intro}
Images captured in low-light environments usually contain large scales of dark areas. They are of low contrast and intensities, which submerge the useful image contents, making them invisible to humans as well as damaging the performances of numerous computer vision algorithms \cite{huang-cvpr20-real,li-iccvw21-photon,liu-ijcv21-benchmarking,xu-tmcca21-exploring}. Many attempts have been made to solve the Low-Light Image Enhancement (LLIE) task. Besides the early methods, e.g., histogram equalization \cite{pizer-vbc1990-contrast}, Retinex-based optimization methods \cite{guo-tip16-lime,ng-jis11-a,ren-tip20-lr3m,li-tip18-structure,hao-tmm20-low}, CNNs are introduced to learn illumination-recovering deep representations from low-light images. Many CNN-based methods~\cite{zhang-arxiv20-self,zhang-acmmm19-kindling,zhang-ijcv21-beyond,wu-cvpr22-uretinex,wei-bmvc18-deep} solve LLIE by combining Retinex theory~\cite{jobson1997properties}, which decomposes an image $\mathbf{I}$ into reflectance $\mathbf{R}$ and illumination $\mathbf{L}$:
\begin{equation}\label{eq:retinex}
\mathbf{I} =\mathbf{R} \odot \mathbf{L}
\end{equation}
where $\odot$ is point-wise multiplication. There still exist some other deep networks learning a mapping from low-light images to normal-light ones, e.g., the supervised methods \cite{chen-cvpr18-learning,lore-pr17-a,kim-iccv21-representative,yang-cvpr22-adaint,koh-cvpr22-bnudc,zhang-cvpr22-deep,dong-cvpr22-abandoning}, and unsupervised methods \cite{guo-cvpr20-zero,jiang-tip21-enlightengan,ma-cvpr22-toward}.

\begin{figure}[t]
\begin{center}
\begin{minipage}{0.32\linewidth}
\centering{\includegraphics[width=1\linewidth]{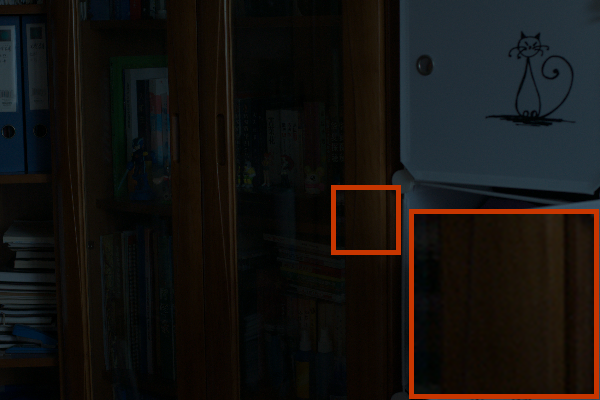}}
\centerline{Input}
\end{minipage}
\begin{minipage}{0.32\linewidth}
\centering{\includegraphics[width=1\linewidth]{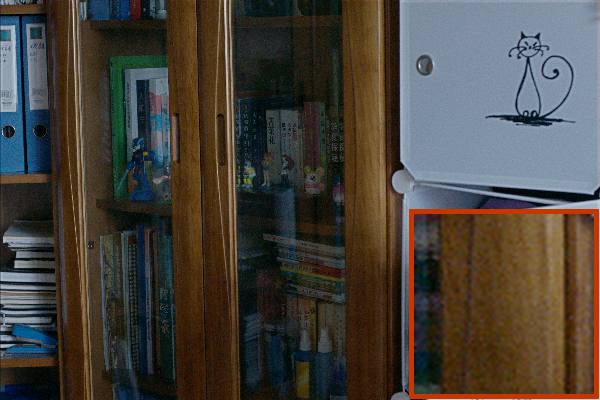}}
\centerline{SCI \cite{ma-cvpr22-toward}}
\end{minipage}
\begin{minipage}{0.32\linewidth}
\centering{\includegraphics[width=1\linewidth]{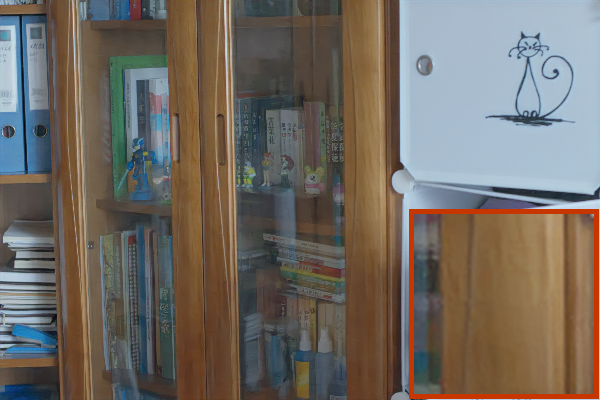}}
\centerline{URetinexNet \cite{wu-cvpr22-uretinex}}
\end{minipage}
\vfill
\begin{minipage}{0.32\linewidth}
\centering{\includegraphics[width=1\linewidth]{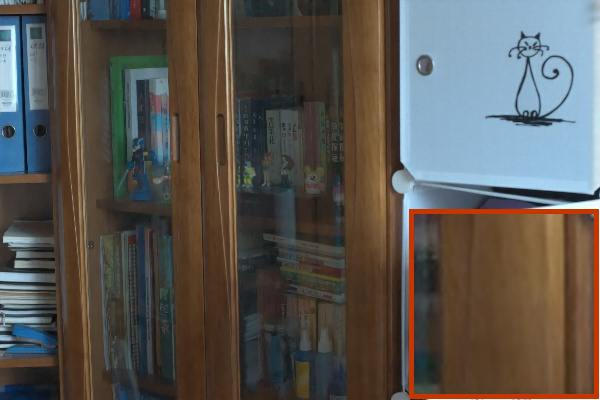}}
\centerline{SNR \cite{xu-cvpr22-snr}}
\end{minipage}
\begin{minipage}{0.32\linewidth}
\centering{\includegraphics[width=1\linewidth]{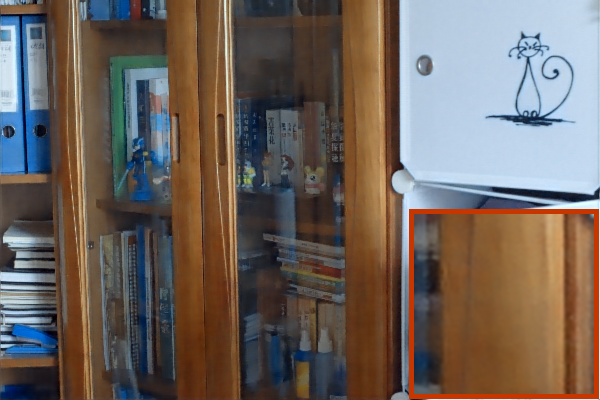}}
\centerline{Ours}
\end{minipage}
\begin{minipage}{0.32\linewidth}
\centering{\includegraphics[width=1\linewidth]{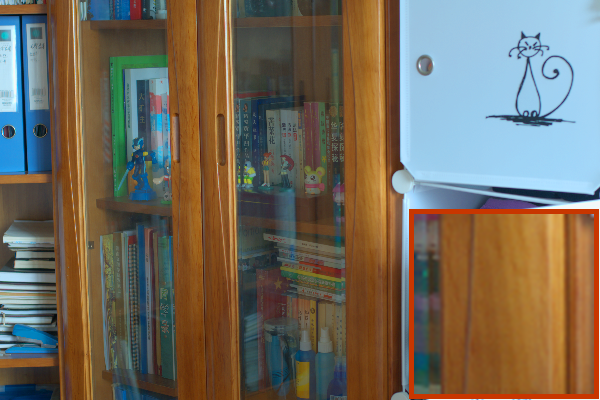}}
\centerline{GT}
\end{minipage}
\end{center}
\caption{This figure shows a challenging low-light image, whose middle areas are completely swallowed by the dark. Our IAGC obtains higher illumination and recovers more image details than SOTA methods. This is attributed to illumination-aware gamma correction which effectively enhances illumination as well as COMO-ViT which recovers better image details by completely modelling the dependencies among all pixels of images.}
\label{fig:teaser}
\end{figure}

However, the convolutional mechanism is usually limited by its inductive biases~\cite{naseer2021intriguing}. The locality is unable to handle long-range dark areas well, and the spatial invariance regards dark and bright areas in the same manner, so that some useful information in the bright areas cannot be properly used. Therefore, CNN-based methods are insensitive to learning effective illumination-recovering deep representations, leading to still low illumination and inaccurate color recovery for some images.

Transformers are introduced to enhance low-light images by modelling longer-range context dependencies \cite{zhang-iccv21-star,xu-cvpr22-snr}. An effective paradigm is to combine self-attention and convolution in a complementary way. However, these methods usually downsample an image before calculating self-attention to reduce computation. The dependencies among some pixels are inevitably ignored, losing some useful information and leading to unsatisfactory recovery results for some low-light images.

Allowing for the insensitivity of CNNs to handling low-light images, we propose to make the best use of the effectiveness of Gamma Correction to illumination enhancement. Existing methods usually neglect the differences between different image contents and even different images by using single empirical gamma values \cite{guo-tip16-lime,hao-tmm20-low,li-tip18-structure,ng-jis11-a,ren-tip20-lr3m}. We propose to adaptively learn correction factors for different image contents according to their unique brightness by introducing an Illumination-Aware Gamma Correction network (IAGC). IAGC integrates the effectiveness of gamma correction with the strong modelling capacities of deep neural networks, so that more effective gamma can be evaluated via perceiving different image contents. 

IAGC adopts a coarse-to-fine strategy to learn correction factors. Coarsely, we globally recover the brightness of low-light images by learning a whole correction factor with a Global Gamma Correction Module (GGCM) perceiving global illumination. Finely, we further build a Local Gamma Correction Module (LGCM) to locally explore pixel dependencies to elaborately enhance illumination according to different local image contents. An example is shown in Fig.~\ref{fig:teaser} to illustrate our improvement on visual quality over existing state-of-the-art LLIE works. 

In our work, gamma is learnable, gamma correction thus introduces exponential operation in both forward and backward propagation, bringing high computational complexity. We propose to use Taylor Series to approximate gamma correction so that exponential operations are avoided in both forward and backward propagation.

Low-light images usually contain large-scale dark areas, which are ineffective to be enhanced by CNNs due to their locality bias. Similarly, calculating attention patch-wisely \cite{liang-iccvw21-swinir} or on a downsampled image \cite{xu-cvpr22-snr,zhang-iccv21-star} is also verified to be not enough to handle such large dark areas. We propose to completely model the dependencies of all pixels in an image, so that information from far bright regions can be used to recover dark areas. To reach such a goal, we propose a COmpletely MOdelling Vision Transformer (COMO-ViT) block to describe images via a local-to-global hierarchical self-attention mechanism. Extensive experiments in later sections verify the stronger modelling capacity of our COMO-ViT for modelling long-range image dependencies than state-of-the-art Transformer blocks.

The main contributions of this paper are summarized as:
\begin{itemize}
    \setlength{\itemsep}{2pt}
    \setlength{\parsep}{2pt}
    \setlength{\parskip}{2pt}
    \item We are the first to propose utilizing learnable gamma correction to adaptively solve the insensitivity of existing models to extract effective illumination-recovering deep representation. To reach this goal, two illumination-aware gamma correction modules (GGCM and LGCM) are introduced to predict global and local adaptive correction factors. Moreover, Taylor Series are used to approximate exponential operation to improve running efficiency.
    \item We propose to completely express images by exploring extensive dependencies of all pixels. A novel vision Transformer (COMO-ViT) block is thus introduced to connect all pixels via a local-to-global hierarchical self-attention. COMO-ViT is verified to be able to extract highly-effective deep features and keep strong modelling capacities.
    \item We introduce an effective network architecture, i.e., IAGC, for low-light image enhancement by combining GGCM, LGCM, and COMO-ViT block together. Extensive experiments show that we obtain better quantitative evaluation and hue recovery than SOTA methods compared with ground truth.
\end{itemize}

\begin{figure*}[t]
\begin{center}
\begin{minipage}{0.95\linewidth}
\centering{\includegraphics[width=1\linewidth]{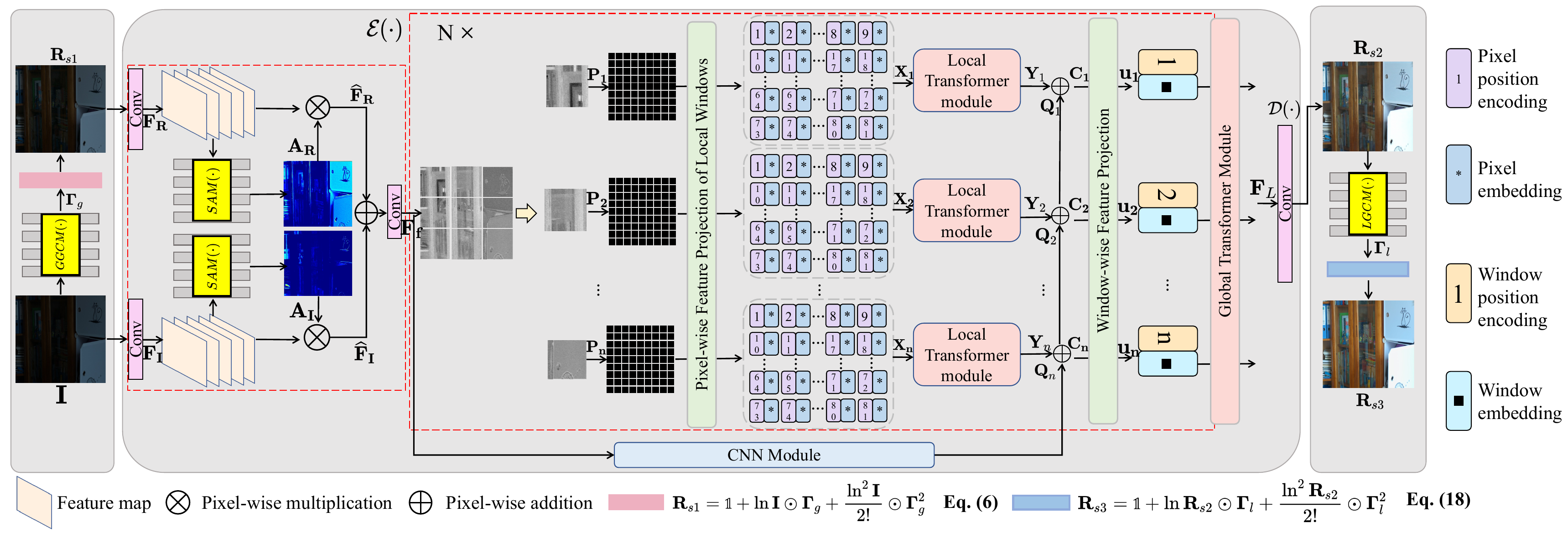}}
%\centerline{(a)}
\end{minipage}
\end{center}
\caption{Overview of our IAGC framework. We adopt a three-stage illumination enhancement strategy. Firstly, we coarsely enhance $\mathbf{I}$ via a global gamma correction module (GGCM). Secondly, an encoder-decoder structure is used to further enhance illumination and remove noise. Finally, we introduce another local gamma correction module (LGCM) to elaborately tune the illumination.}
\label{fig:network}
\end{figure*}
%-------------------------------------------------------------------------
\section{Related Work}
In this section, we comprehensively review existing works on low-light image enhancement and analyze their limitations in producing satisfactory results.

\vspace{-4mm}
\paragraph{Conventional Methods.} Many conventional LLIE methods are based on Retinex theory, which formulates an image as a multiplication of reflectance and illumination (Eq.~\eqref{eq:retinex}). Some methods estimate the reflectance and illumination via a variational framework, the estimated illumination is further adjusted to restore the original low-light image \cite{fu-cvpr16-a,kimmel-ijcv03-a,ng-jis11-a}. Other Retinex-based methods usually optimize an energy function derived from the Maximum-a-Posteriori (MAP) framework to enhance low-light images, in which some heuristic priors are designed to constrain the properties of reflectance or illumination \cite{guo-tip16-lime,hao-tmm20-low,li-tip18-structure,ren-tip20-lr3m}. For example, Gaussian total variation in \cite{hao-tmm20-low}, structure-aware regularization in \cite{guo-tip16-lime} and the additional noise term in \cite{li-tip18-structure}. However, these hand-craft priors are usually inaccurate and limited when applied to different cases, leading to apparent color deviation.

% \vspace{-4mm}
\paragraph{CNN-based Learning Methods.} CNN has pushed forward the solution of LLIE by large margins \cite{li-pami21-low}. Some methods integrate Retinex theory with CNNs \cite{wei-bmvc18-deep,zhang-arxiv20-self,zhang-acmmm19-kindling,zhang-ijcv21-beyond,wu-cvpr22-uretinex,wei-bmvc18-deep}. For example, Wei \emph{et al.} propose an end-to-end Retinex-Net, in which a decomposition module and an illumination adjustment module are used to learn reflectance and illumination \cite{wei-bmvc18-deep}. Wu \emph{et al.} propose URetinex-Net which unfolds an optimization problem into several learnable network modules \cite{wu-cvpr22-uretinex}. Some methods design effective networks to predict normal-light images directly from low-light ones via a supervised learning strategy \cite{chen-cvpr18-learning,lore-pr17-a,kim-iccv21-representative,yang-cvpr22-adaint,koh-cvpr22-bnudc,zhang-cvpr22-deep,dong-cvpr22-abandoning}. For instance, \cite{yang-cvpr22-adaint} learns a 3D lookup table, and \cite{zhang-cvpr22-deep} uses color consistency to constrain network training.  There are still some  unsupervised methods to solve LLIE \cite{guo-cvpr20-zero,jiang-tip21-enlightengan,ma-cvpr22-toward}. One famous work is \cite{ma-cvpr22-toward}, in which a self-calibrated illumination learning framework is proposed for fast, flexible, and robust low-light image enhancement. Because CNNs are usually limited by inductive bias, the illumination still cannot be well recovered for some images.

\vspace{-4mm}
\paragraph{Transformer-based Methods.} Transformer-based methods usually combine self-attention and convolution to extract long- and short-range dependencies, so that better performance can be obtained \cite{xu-cvpr22-snr,zhang-iccv21-star,wang-cvpr23-smartassign}. However, the long-range context dependencies are usually obtained by building patch-level relationships, and these methods downsample images before calculating self-attention to reduce computation. Hence, they may lose some useful information, which is adverse to recover high-quality images.

In this paper, we propose a local-to-global hierarchical self-attention to completely express images. We would like to point out that the work \cite{han-nips21-transformer} aiming at high-level tasks proposes a TNT structure to explore local attention to assist global attention. Our work is different from it in the following aspects: 1) TNT still focuses on patch-level attention, pixel-level dependencies are still not explored, i.e., it does not completely model images. While ours completely express images via pixel-level attention. 2) Local attention in TNT just provides auxiliary information for the patch embedding of global attention. While our local attention models pixel dependencies locally and the global attention extends pixel dependencies to the whole image. 3) TNT adopts a bypass structure to realize auxiliary information enhancement. While our local attentions are directly embedded as the input for global attention.
 
%-------------------------------------------------------------------------
\section{Our Method}
\label{sec:method}
We first overview the proposed IAGC in Sec.~\ref{sec:overview}. Then, we present its architecture with illumination-aware gamma correction modules (GGCM and LGCM) and the completely modelling Transformer (COMO-ViT) block in Sec.~\ref{sec:arch}. Finally, we introduce loss functions used to train IAGC in Sec.~\ref{sec:loss}.

\subsection{Overview}
\label{sec:overview}
\begin{figure}[t]
\begin{center}
\begin{minipage}{0.45\linewidth}
\centering{\includegraphics[width=1\linewidth]{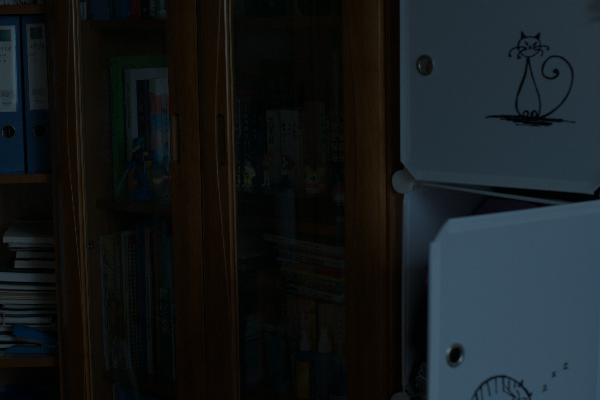}}
\centerline{$\mathbf{I}$}
\end{minipage}
\begin{minipage}{0.45\linewidth}
\centering{\includegraphics[width=1\linewidth]{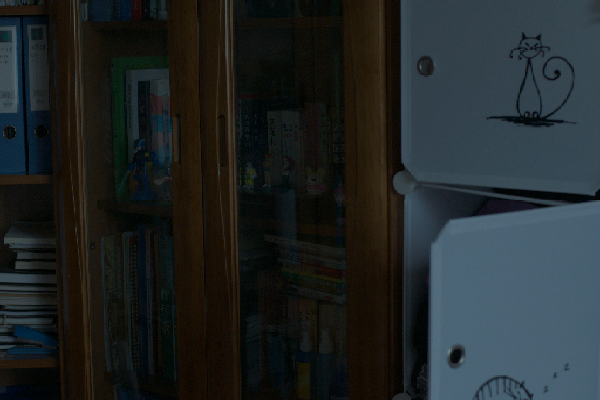}}
\centerline{$\mathbf{R}_{s1}$}
\end{minipage}
\vfill
\begin{minipage}{0.45\linewidth}
\centering{\includegraphics[width=1\linewidth]{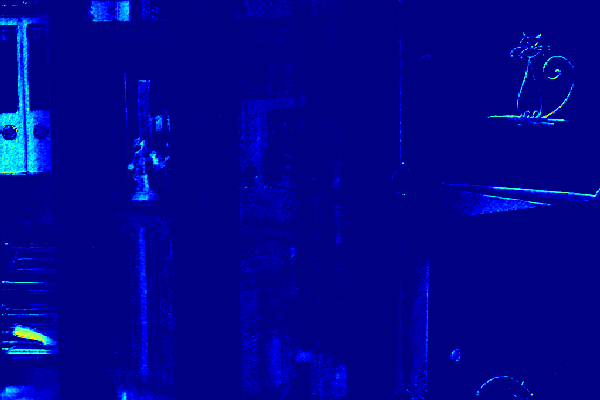}}
\centerline{$\mathbf{A}_{\mathbf{I}}$}
\end{minipage}
\begin{minipage}{0.45\linewidth}
\centering{\includegraphics[width=1\linewidth]{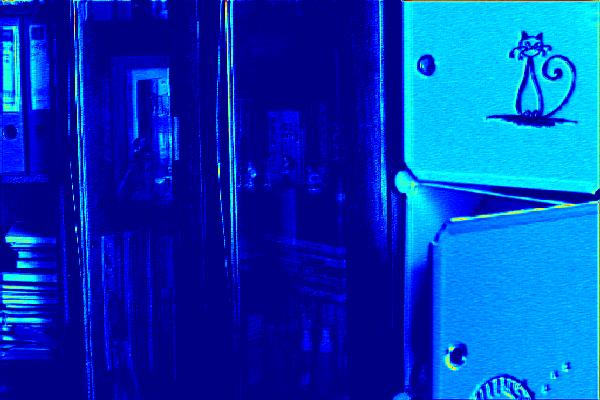}}
\centerline{$\mathbf{A}_{\mathbf{R}}$}
\end{minipage}
\end{center}
%\vspace{4mm}
\caption{This figure shows that GGCM enables SAM to capture more informative image contents, which provides more knowledge for feature learning at the next stage. The learned correction factor $\mathbf{\Gamma}_{g}=0.5976$ for this example.}
\label{fig:exam}
\end{figure}

The overall architecture of IAGC is illustrated in Fig.~\ref{fig:network}. Given a low-light image $\mathbf{I}$, we propose to enhance its illumination via a three-stage coarse-to-fine strategy. In the first stage, we introduce a global gamma correction module (GGCM) to extract illumination features and learn a whole correction factor $\mathbf{\Gamma}_{g}$ to coarsely enhance $\mathbf{I}$:
\begin{equation}\label{eq:ggcm}
\begin{aligned}
\mathbf{\Gamma}_{g} = \operatorname{GGCM}(\mathbf{I}), \quad
\mathbf{R}_{s1} = \mathbf{I}^{\mathbf{\Gamma}_{g}},
 \end{aligned}
\end{equation}
where $\mathbf{R}_{s1}$ is the enhanced result. Compared with $\mathbf{I}$, $\mathbf{R}_{s1}$ is of higher brightness and many informative contents emerge from the dark areas. Experiments also show that GGCM enables attention to capture more informative image contents, as shown in Fig.~\ref{fig:exam}, providing more knowledge for the second stage. Therefore, our GGCM is effective to solve the insensitivity of existing methods to enhance illumination and recover the swallowed contents in the low-light image $\mathbf{I}$. 

In the second stage, we fuse the information in  $\mathbf{I}$ and $\mathbf{R}_{s1}$ and build a backbone encoder $\mathcal{E}(\cdot)$ to learn illumination-recovered representations $\mathbf{F}$ from both of them, and then further enhance brightness and recover image details via a decoder $\mathcal{D}(\cdot)$. Because low-light images usually contain large-scale dark areas and only a small part may possess relatively higher illumination, it is important to completely model pixel-level dependencies of images to infer the contents in the dark based on the informative bright areas. We introduce a novel vision Transformer block COMO-ViT in $\mathcal{E}(\cdot)$ to connect all pixels via a local-to-global self-attention. Due to the strong modelling capacity of COMO-ViT, the noise which may appear in the illumination enhancement \cite{xu-cvpr22-snr} is also implicitly removed during this stage, as shown in Fig.~\ref{fig:exam_result}. We formulate this stage as:
\begin{equation}\label{eq:stage2}
\begin{aligned}
\mathbf{F} = \mathcal{E}(\mathbf{I}, \mathbf{R}_{s1}), \quad 
\mathbf{R}_{s2} = \mathcal{D}(\mathbf{F}),
 \end{aligned}
\end{equation}
where $\mathbf{R}_{s2}$ is the enhanced result of the second stage. 

In order to obtain better hue recovery compared with ground truth, we introduce a local gamma correction module (LGCM) to further perceive the illumination gaps between $\mathbf{R}_{s2}$ and the ground truth. The pixel-wise correction factor $\mathbf{\Gamma}_{l}$ is predicted to enhance illumination in a more elaborate manner, which is formulated as:
\begin{equation}\label{eq:lgcm}
\begin{aligned}
\mathbf{\Gamma}_{l} = \operatorname{LGCM}(\mathbf{R}_{s2}), \quad
\mathbf{R}_{s3} = \mathbf{R}_{s2}^{\mathbf{\Gamma}_{l}},
 \end{aligned}
\end{equation}
where $\mathbf{R}_{s3}$ is our final low-light enhancement result.

Different from existing models learning illumination-recovered representation directly from the less informative low-light images, we introduce GGCM and LGCM to perceive illumination differences and adaptively predict correction factors, so that the degraded illumination and image details are effectively recovered.  Complementarily, our COMO-ViT explores better pixel dependencies of an image via a local-to-global hierarchical self-attention. Hence, our IAGC makes better use of the bright pixels to recover every dark pixel than existing Transformer-based models which only develop patch-level dependencies \cite{xu-cvpr22-snr,zhang-iccv21-star}.

\subsection{Architecture of IAGC}
\label{sec:arch}
In this section, we give the detailed structure of our IAGC and meanwhile show some intermediate results to illustrate the effectiveness of our key designs.

\vspace{-4mm}
\paragraph{GGCM.} GGCM takes $\mathbf{I}$ as input, perceives its illumination, and predicts a whole correction factor $\mathbf{\Gamma}_{g}$ via exploring global deep features. Its architecture is formulated as:
\begin{equation}\label{eq:ggcm_formu}
\begin{aligned}
\mathbf{\Gamma}_{g} = \phi(\operatorname{FC}(\operatorname{Avg-Pool}(\operatorname{Conv}(\mathbf{I})))), \quad
\mathbf{R}_{s1} = \mathbf{I}^{\mathbf{\Gamma}_{g}},
\end{aligned}
\end{equation}
where $\operatorname{Conv}(\cdot)$ is a convolutional layer projecting $\mathbf{I}$ into feature space, $\operatorname{Avg-Pool}(\cdot)$ is an 
average pooling layer which extracts global features by squeezing spatial dimensions, $\operatorname{FC}(\cdot)$ is a fully-connected layer for fusing channel information, and $\phi(\cdot)$ is a Sigmoid layer to predict $\mathbf{\Gamma}_{g}$. For simplicity, we omit the operation expanding a scalar to a matrix $\mathbf{\Gamma}_{g}$. Note that Eq. \eqref{eq:ggcm_formu} introduces exponential operation with high complexity during both forward and backward propagation, we utilize Taylor Series to approximate it:
\begin{equation}\label{eq:ggcm_taylor}
\begin{aligned}
\mathbf{R}_{s1} = \mathbbm{1} + \ln\mathbf{I} \odot \mathbf{\Gamma}_{g} + \frac{\ln^{2}\mathbf{I}}{2!} \odot \mathbf{\Gamma}^{2}_{g}
\end{aligned}
\end{equation}
where $\mathbbm{1}$ is all-1 matrix. Fig.~\ref{fig:exam} visualizes an example of $\mathbf{I}$ and corresponding $\mathbf{R}_{s1}$, showing that more image details emerge in $\mathbf{R}_{s1}$, especially at the middle regions.

We then fuse $\mathbf{I}$ and $\mathbf{R}_{s1}$ to supply more knowledge for the second stage by projecting them into feature space:
\begin{equation}\label{eq:conv_proj}
 \mathbf{F}_{\mathbf{I}} = \operatorname{Conv}(\mathbf{I}), \quad \mathbf{F}_{\mathbf{R}} = \operatorname{Conv}(\mathbf{R}_{s1}).
\end{equation}
Because bright pixels are more informative and play a more important role in illumination enhancement than dark ones, we adopt a \textbf{S}patial \textbf{A}ttention \textbf{M}odule (SAM) to highlight bright areas by large weights and constrain dark pixels to reduce their negative effect by smaller weights:
\begin{equation}\label{eq:sam}
\begin{aligned}
 \mathbf{A}_{\mathbf{I}} &= \phi(\operatorname{Conv}(\mathbf{F}_{\mathbf{I}})), \quad  \widehat{\mathbf{F}}_{\mathbf{I}} = \mathbf{A}_{\mathbf{I}} \odot \mathbf{F}_{\mathbf{I}}, \\
 \mathbf{A}_{\mathbf{R}} &= \phi(\operatorname{Conv}(\mathbf{F}_{\mathbf{R}})), \quad  \widehat{\mathbf{F}}_{\mathbf{R}} = \mathbf{A}_{\mathbf{R}} \odot \mathbf{F}_{\mathbf{R}},
\end{aligned}
\end{equation}
where $\mathbf{A}_{\mathbf{I}}$ and $\mathbf{A}_{\mathbf{R}}$ are the spatial attention map of $\mathbf{F}_{\mathbf{I}}$ and $\mathbf{F}_{\mathbf{R}}$, respectively. In Fig.~\ref{fig:exam}, we visualize $\mathbf{A}_{\mathbf{I}}$ and $\mathbf{A}_{\mathbf{R}}$, observing that $\mathbf{A}_{\mathbf{R}}$ captures more informative image contents, which further illustrates the effectiveness of our GGCM. Finally, the fused deep feature $\mathbf{F}_{f}$ is obtained as:
\begin{equation}\label{eq:fuse}
 \mathbf{F}_{f} = \operatorname{Conv}(\widehat{\mathbf{F}}_{\mathbf{I}}+\widehat{\mathbf{F}}_{\mathbf{R}}).
\end{equation}

\begin{table*}[t]
\begin{center}
\begin{tabular}{cccccccccccc}
\hline
\hline
Epoches & Optimizer & \makecell[c]{Batch \\ size} & \makecell[c]{Learning \\ rate} & \makecell[c]{LR \\ decay} & \makecell[c]{Weight \\ decay} & \makecell[c]{Drop \\ path} & \makecell[c]{Embedding \\ dim} & Head & $L$ & \makecell[c]{Window \\ size} & \makecell[c]{Patch \\ size} \\
\hline
300 & Adam\cite{kingma-iclr15-adam} & 8 & 4e-4 & cosine & 1e-7 & $0.1$ & $15$ & $5$ & $2$ & $16$ & 512\\
\hline
\hline
\end{tabular}
\end{center}
\caption{Default training and network hyper-parameters used in our method, unless stated otherwise.}
\label{tab:hyper_para}
\end{table*}

\vspace{-4mm}
\paragraph{COMO-ViT.} Given the input feature $\mathbf{F}_{l-1} \in \mathbb{R}^{H \times W \times c}$ of COMO-ViT, we conduct two branches of operations. In the first branch, we uniformly split it into $n$ non-overlapping windows $\mathcal{P} = [\mathbf{P}^1, \mathbf{P}^2, \cdots, \mathbf{P}^n] \in \mathbb{R}^{n \times w \times w \times c}$, where $(w, w)$ is the window resolution. SNR \cite{xu-cvpr22-snr} and STAR \cite{zhang-iccv21-star} downsample images, losing local structures and some important pixel-level information. Instead, the proposed COMO-ViT completely models the dependencies among all pixels of an image via a local-to-global hierarchical self-attention. Locally, each pixel in a window $\mathbf{P}^{i}$ is regarded as an individual, we thus reshape $\mathbf{P}^{i}$ as follows:
\begin{equation}\label{eq:reshape_window}
\begin{aligned}
\mathbf{P}^{i} \to [p^{i,1}, p^{i,2}, \cdots, p^{i,m}],
\end{aligned}
\end{equation}
where $p^{i,j}\in \mathbb{R}^{1\times 1 \times c}$, $m=w^2$ is the number of pixels in $\mathbf{P}^{i}$. With a linear projection, we then transform the pixels into a sequence of pixel embeddings $\mathbf{X}^{i} = [x^{i,1}, x^{i,2}, \cdots, x^{i,m}]$, where $x^{i,j}\in \mathbb{R}^{c1}$ is the $j$-th pixel embedding, $c1$ is the embedding dimension. For $\mathbf{X}^{i}$, we utilize a local Transformer module to extract deep features as follows:
\begin{equation}\label{eq:local_trans}
\begin{aligned}
{\mathbf{Y}^{'}}^{i} &= \mathbf{X}^{i} + \operatorname{MSA}(\operatorname{LN}(\mathbf{X}^{i})), \\
\mathbf{Y}^{i} &= {\mathbf{Y}^{'}}^{i} + \operatorname{MLP}(\operatorname{LN}({\mathbf{Y}^{'}}^{i})),
\end{aligned}
\end{equation}
where $\mathbf{Y}^{i}$ is the feature learned by the local Transformer module, $\operatorname{MSA}(\cdot)$ is the Multi-head Self-Attention \cite{vaswani-nips17-attention}, $\operatorname{LN}(\cdot)$ is layer normalization \cite{ba-arxiv16-layer} for stable training and faster convergence, $\operatorname{MLP}(\cdot)$ is multi-layer perceptron for feature transformation at channel dimension and non-linearity. In such a process, we adopt 1D learnable location embedding to encode the spatial information of pixels.

To complement the non-overlapping window attention, in the second branch which is parallel with local attention, we use a CNN module to model local pixel dependencies in $\mathbf{F}_{l-1}$ via an overlapped sliding kernel to recover image details, in which a SE block \cite{hu-cvpr18-squeeze} is used to explore channel relationship to boost representative power:
\begin{equation}\label{eq:local_cnn}
\begin{aligned}
\mathbf{F}^{'} = \operatorname{Conv}(\operatorname{LN}(\mathbf{F}_{l-1})), \quad
\mathbf{F}_{conv} = \mathbf{F}^{'} \odot \operatorname{SE}(\mathbf{F}^{'}).
\end{aligned}
\end{equation}
$\mathbf{F}_{conv}$ is then split into $n$ non-overlapping windows $\mathcal{Q} = [\mathbf{Q}^1, \mathbf{Q}^2, \cdots, \mathbf{Q}^n] \in \mathbb{R}^{n \times w \times w \times c}$, and each $\mathbf{Q}^{i}$ is reshaped:
\begin{equation}\label{eq:reshape_windowq}
\begin{aligned}
\mathbf{Q}^{i} \to [q^{i,1}, q^{i,2}, \cdots, q^{i,m}].
\end{aligned}
\end{equation}
We combine the features from both branches as:
\begin{equation}\label{eq:comb_fea}
\begin{aligned}
\mathcal{C} = [\mathbf{C}^1, \mathbf{C}^2, \cdots, \mathbf{C}^n], \quad \mathbf{C}^{i} = \mathbf{Q}^{i} + \mathbf{Y}^{i}
\end{aligned}
\end{equation}

Global pixel dependencies are explored by calculating window attention via a global attention module. Firstly, $\mathcal{C}$ is transformed into a sequence of window embedding:
\begin{equation}\label{eq:win_embed}
\begin{aligned}
\mathbf{U} = [u^1, u^2, \cdots, u^n], \quad u^{i} = \operatorname{FC}(\operatorname{Vec}(\mathbf{C}^{i})),
\end{aligned}
\end{equation}
where $\operatorname{Vec}(\cdot)$ is vectorization operation. Then, we utilize a global Transformer module to explore inter-window dependencies, obtaining the feature $\mathbf{F}_{l} \in \mathbb{R}^{H \times W \times c}$, where $l\in \{1, 2, \cdots, L\}$, and $L$ is the COMO-ViT number. When $l=1$, $\mathbf{F}_{l-1}$ is the fused feature $\mathbf{F}_{f}$ in Eq.~\eqref{eq:fuse}.

The result of the second stage is obtained by decoding $\mathbf{F}_{L}$ with a convolutional layer ($\mathcal{D}(\cdot)$):
\begin{equation}\label{eq:out_stage2}
\begin{aligned}
\mathbf{R}_{s2} = \operatorname{Conv}(\mathbf{F}_{L}).
\end{aligned}
\end{equation}

\begin{figure}[t]
\begin{center}
\begin{minipage}{0.45\linewidth}
\centering{\includegraphics[width=1\linewidth]{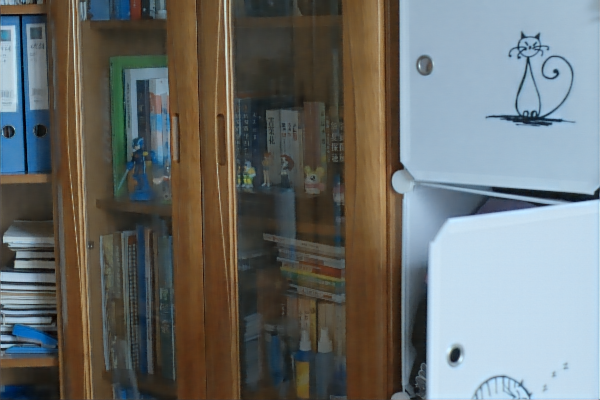}}
\centerline{$\mathbf{R}_{s2}$}
\end{minipage}
\begin{minipage}{0.45\linewidth}
\centering{\includegraphics[width=1\linewidth]{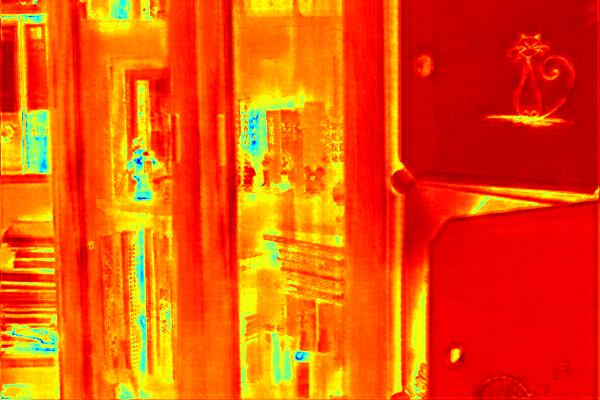}}
\centerline{$\mathbf{\Gamma}_{l}$}
\end{minipage}
\begin{minipage}{0.45\linewidth}
\centering{\includegraphics[width=1\linewidth]{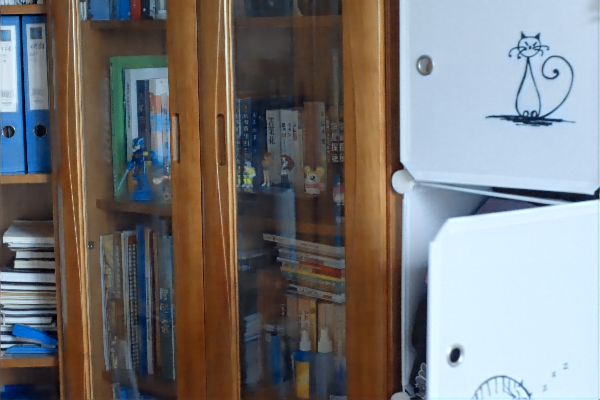}}
\centerline{$\mathbf{R}_{s3}$}
\end{minipage}
\begin{minipage}{0.45\linewidth}
\centering{\includegraphics[width=1\linewidth]{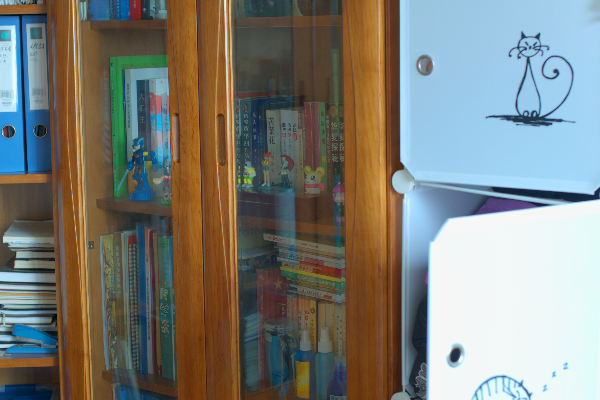}}
\centerline{GT}
\end{minipage}
\end{center}
\caption{We observe that $\mathbf{R}_{s3}$ keeps higher illumination than $\mathbf{R}_{s2}$, illustrating the effectiveness of our LGCM. $\mathbf{\Gamma}_{l}$ is corresponding pixel-wise local gamma map.}
\label{fig:exam_result}
\end{figure}

\begin{figure*}[t]
\begin{center}
\begin{minipage}{0.12\linewidth}
\centering{\includegraphics[width=1\linewidth]{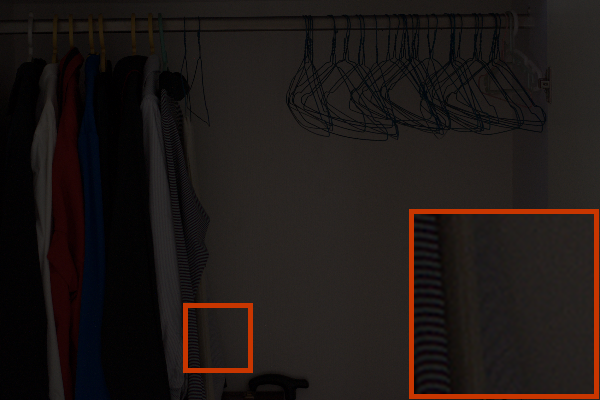}}
%\centerline{(a)}
\end{minipage}
\begin{minipage}{0.12\linewidth}
\centering{\includegraphics[width=1\linewidth]{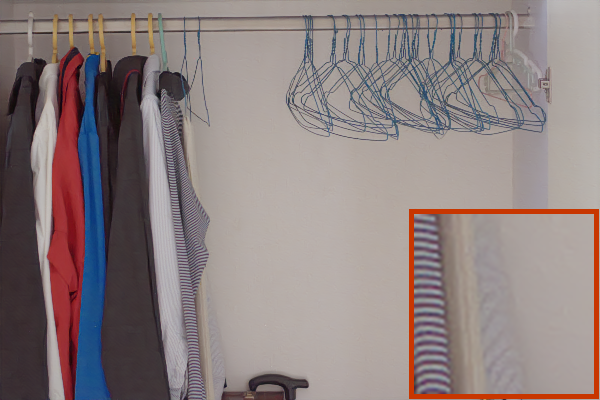}}
%\centerline{(a)}
\end{minipage}
\begin{minipage}{0.12\linewidth}
\centering{\includegraphics[width=1\linewidth]{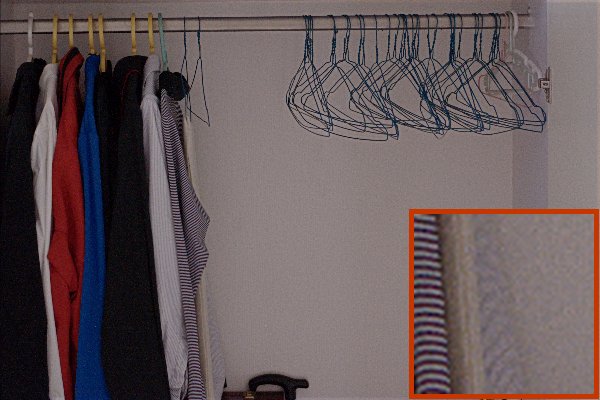}}
%\centerline{(a)}
\end{minipage}
\begin{minipage}{0.12\linewidth}
\centering{\includegraphics[width=1\linewidth]{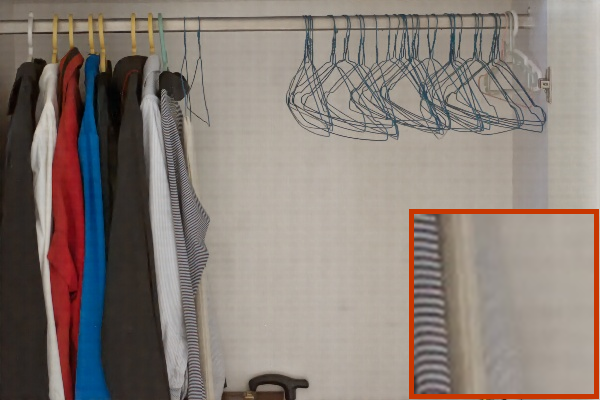}}
%\centerline{(a)}
\end{minipage}
\begin{minipage}{0.12\linewidth}
\centering{\includegraphics[width=1\linewidth]{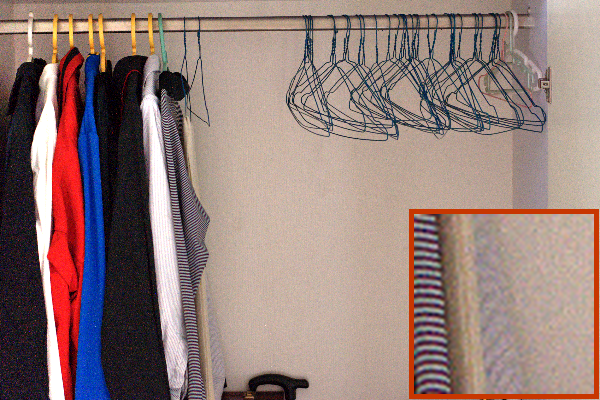}}
%\centerline{(a)}
\end{minipage}
\begin{minipage}{0.12\linewidth}
\centering{\includegraphics[width=1\linewidth]{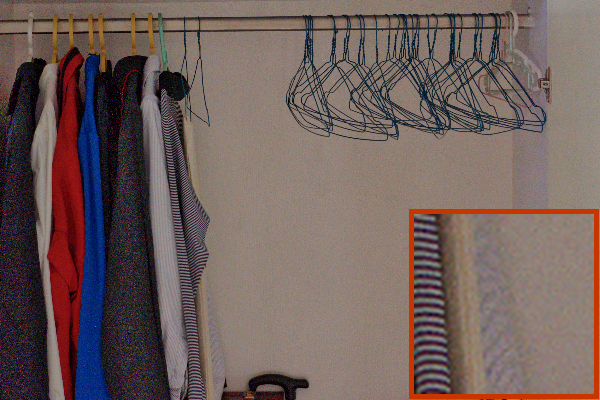}}
%\centerline{(a)}
\end{minipage}
\begin{minipage}{0.12\linewidth}
\centering{\includegraphics[width=1\linewidth]{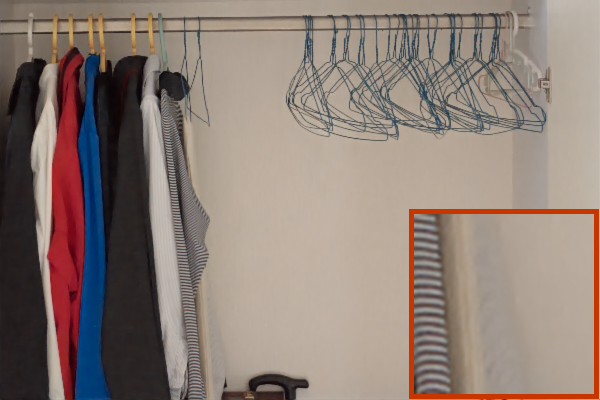}}
%\centerline{(a)}
\end{minipage}
\begin{minipage}{0.12\linewidth}
\centering{\includegraphics[width=1\linewidth]{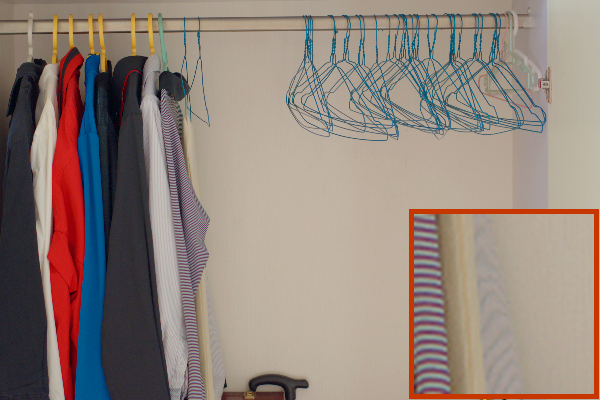}}
%\centerline{(a)}
\end{minipage}
\vfill
\begin{minipage}{0.12\linewidth}
\centering{\includegraphics[width=1\linewidth]{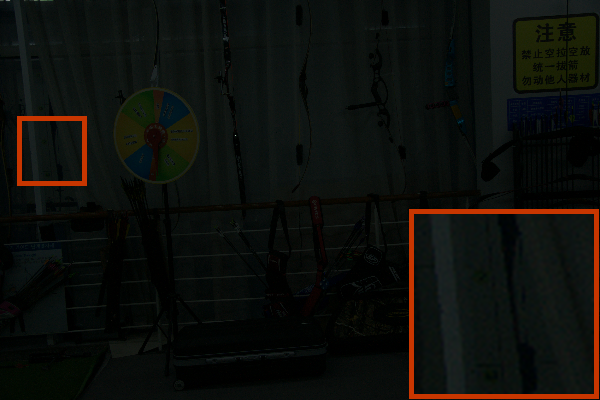}}
%\centerline{(a)}
\end{minipage}
\begin{minipage}{0.12\linewidth}
\centering{\includegraphics[width=1\linewidth]{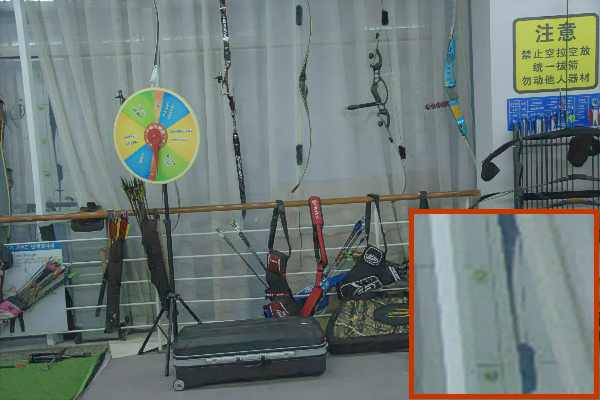}}
%\centerline{(a)}
\end{minipage}
\begin{minipage}{0.12\linewidth}
\centering{\includegraphics[width=1\linewidth]{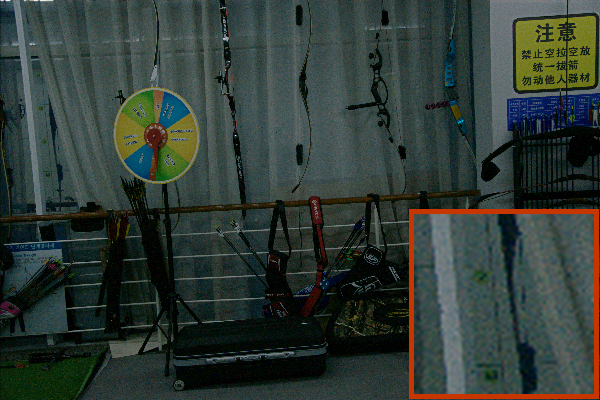}}
%\centerline{(a)}
\end{minipage}
\begin{minipage}{0.12\linewidth}
\centering{\includegraphics[width=1\linewidth]{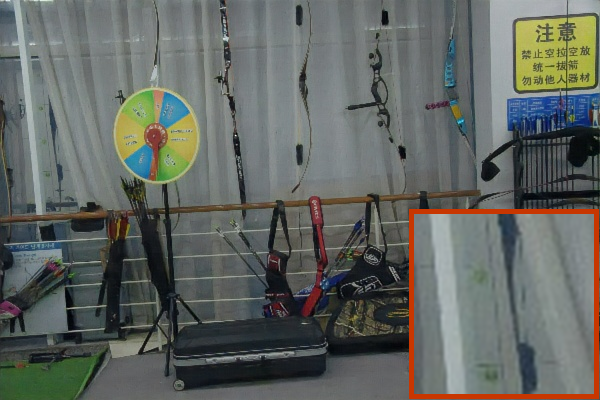}}
%\centerline{(a)}
\end{minipage}
\begin{minipage}{0.12\linewidth}
\centering{\includegraphics[width=1\linewidth]{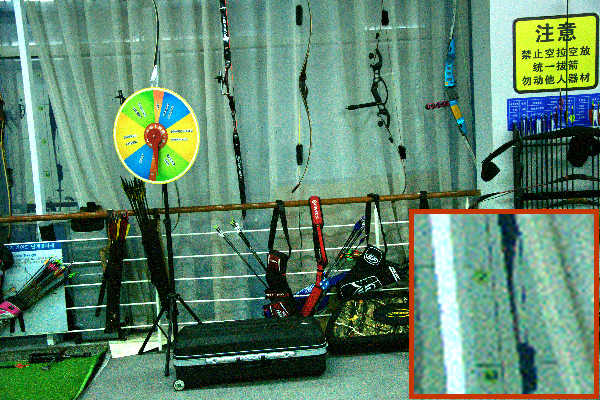}}
%\centerline{(a)}
\end{minipage}
\begin{minipage}{0.12\linewidth}
\centering{\includegraphics[width=1\linewidth]{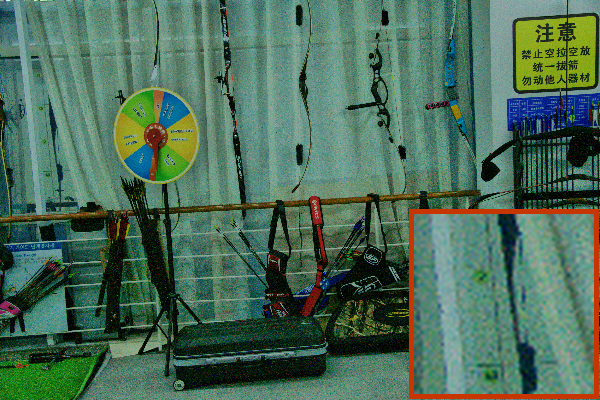}}
%\centerline{(a)}
\end{minipage}
\begin{minipage}{0.12\linewidth}
\centering{\includegraphics[width=1\linewidth]{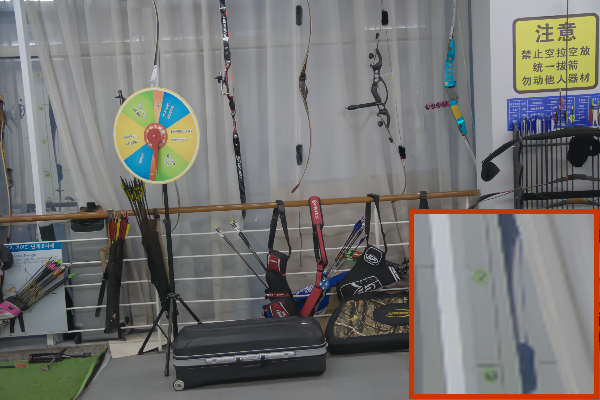}}
%\centerline{(a)}
\end{minipage}
\begin{minipage}{0.12\linewidth}
\centering{\includegraphics[width=1\linewidth]{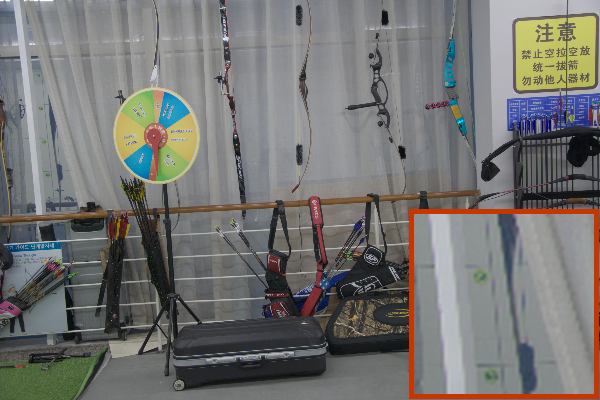}}
%\centerline{(a)}
\end{minipage}
\vfill
\begin{minipage}{0.12\linewidth}
\centering{\includegraphics[width=1\linewidth]{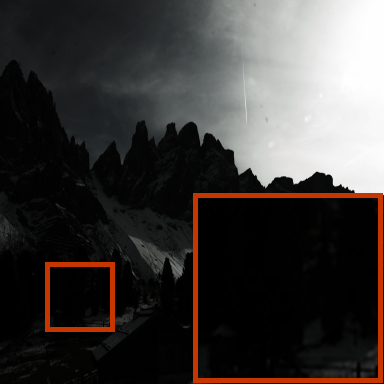}}
\centerline{Input}
\end{minipage}
\begin{minipage}{0.12\linewidth}
\centering{\includegraphics[width=1\linewidth]{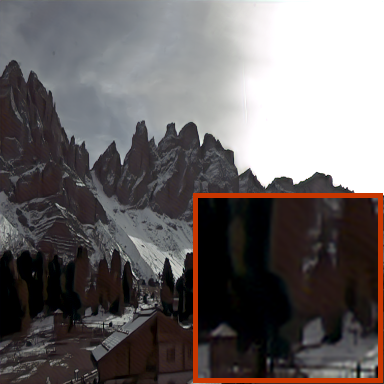}}
\centerline{URetinexNet \cite{wu-cvpr22-uretinex}}
\end{minipage}
\begin{minipage}{0.12\linewidth}
\centering{\includegraphics[width=1\linewidth]{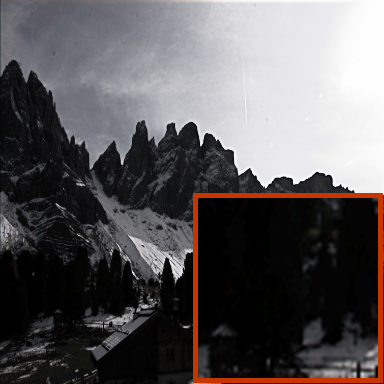}}
\centerline{SCI \cite{ma-cvpr22-toward}}
\end{minipage}
\begin{minipage}{0.12\linewidth}
\centering{\includegraphics[width=1\linewidth]{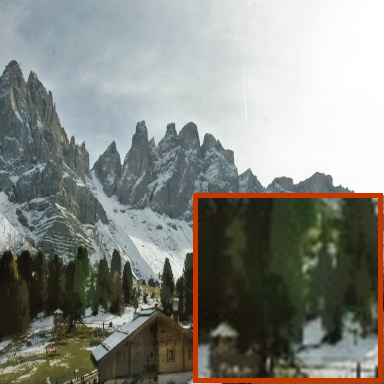}}
\centerline{SNR \cite{xu-cvpr22-snr}}
\end{minipage}
\begin{minipage}{0.12\linewidth}
\centering{\includegraphics[width=1\linewidth]{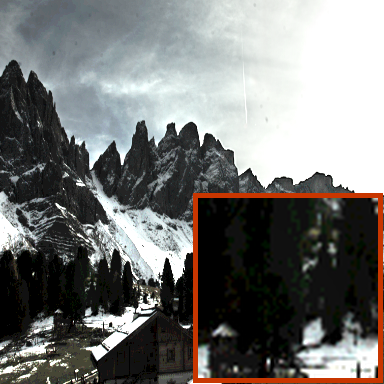}}
\centerline{LIME \cite{guo-tip16-lime}}
\end{minipage}
\begin{minipage}{0.12\linewidth}
\centering{\includegraphics[width=1\linewidth]{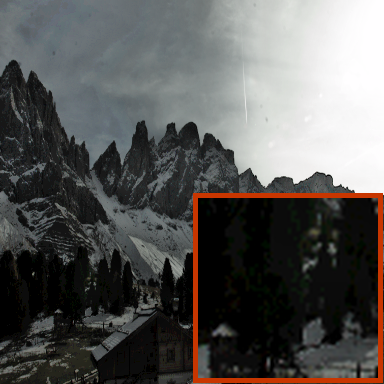}}
\centerline{NPE \cite{wang-tip13-naturalness}}
\end{minipage}
\begin{minipage}{0.12\linewidth}
\centering{\includegraphics[width=1\linewidth]{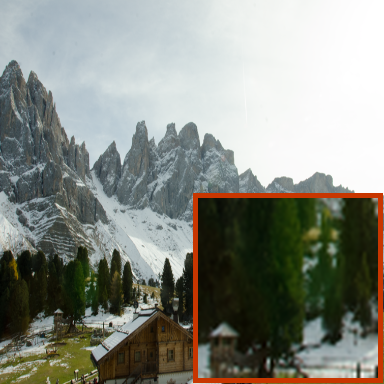}}
\centerline{Ours}
\end{minipage}
\begin{minipage}{0.12\linewidth}
\centering{\includegraphics[width=1\linewidth]{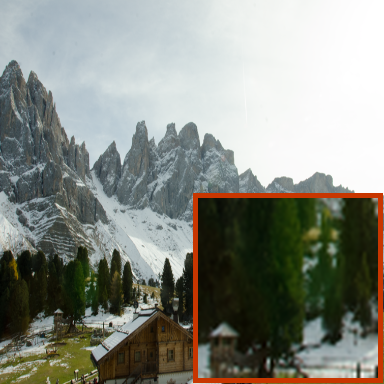}}
\centerline{GT}
\end{minipage}
\end{center}
\caption{Visual quality comparison with SOTA methods on the LOL-v1 (the first row), LOL-v2-real (the second row), and LOL-v2-synthetic (the third row) datasets, respectively. Our method recovers better illumination and preserves more image details.}
\label{fig:real_comp_rain}
\end{figure*}

\renewcommand{\tabcolsep}{5pt}
\begin{table*}[t]
\begin{center}
\begin{tabular}{cccccccccccc}
\hline
\hline
Methods & \makecell[c]{SNR \\ \cite{xu-cvpr22-snr}} & \makecell[c]{URetinex \\ -Net\cite{wu-cvpr22-uretinex}} & \makecell[c]{DeepLPF \\ \cite{moran-cvpr20-deeplpf}} & \makecell[c]{SCI \\ \cite{ma-cvpr22-toward}} & \makecell[c]{LIME \\ \cite{guo-tip16-lime}} & \makecell[c]{MF \\ \cite{fu-sp16-a}} & \makecell[c]{NPE \\ \cite{wang-tip13-naturalness}} & \makecell[c]{SRIE \\ \cite{fu-cvpr16-a}} & \makecell[c]{SDD \\ \cite{hao-tmm20-low}} & \makecell[c]{CDEF \\\cite{lei-tmm22-low}} & \makecell[c]{LPNet \\\cite{li-tmm20-luminance}} \\
\hline
PSNR & 24.49 & 21.32 & 15.28 & 15.80 & 16.76 & 16.96 & 16.96 & 11.86 & 13.34 & 16.33 & 21.46\\
SSIM & 0.840 & 0.835 & 0.473 & 0.527 & 0.444 & 0.505 & 0.481 & 0.493 & 0.635 & 0.583 & 0.802\\
\hline
Methods & \makecell[c]{DRBN \\ \cite{yang-tip21-band}} & \makecell[c]{KinD \\ \cite{zhang-acmmm19-kindling}} & \makecell[c]{RUAS \\ \cite{liu-cvpr21-retinex}} & \makecell[c]{FIDE \\ \cite{xu-cvpr20-learning}} & \makecell[c]{EG \\ \cite{jiang-tip21-enlightengan}} & \makecell[c]{MS-RDN \\ \cite{yang-tip21-sparse}} & \makecell[c]{Retinex \\ -Net\cite{wei-bmvc18-deep}} & \makecell[c]{MIRNet \\ \cite{zamir-eccv20-learning}} & \makecell[c]{IPT \\\cite{chen-cvpr21-pre}} & \makecell[c]{Uformer \\\cite{wang-cvpr22-uformer}} & Ours \\
\hline
PSNR & 20.13 & 20.87 & 18.23 & 18.27 & 17.48 & 17.20 & 16.77 & 24.14 & 16.27 & 16.36 & \textbf{24.53}\\
SSIM & 0.830 & 0.800 & 0.720 & 0.665 & 0.650 & 0.640 & 0.560 & 0.830 & 0.504 & 0.507 & \textbf{0.842}\\
\hline
\hline
\end{tabular}
\end{center}
\caption{Quantitative comparisons with state-of-the-art methods on the LOL-v1 dataset.}
\label{tab:quanti_comp_v1}
\end{table*}

\renewcommand{\tabcolsep}{5pt}
\begin{table*}[t]
\begin{center}
\begin{tabular}{cccccccccccc}
\hline
\hline
Methods & \makecell[c]{SNR \\ \cite{xu-cvpr22-snr}} & \makecell[c]{URetinex \\ -Net\cite{wu-cvpr22-uretinex}} & \makecell[c]{DeepLPF \\ \cite{moran-cvpr20-deeplpf}} & \makecell[c]{SCI \\ \cite{ma-cvpr22-toward}} & \makecell[c]{LIME \\ \cite{guo-tip16-lime}} & \makecell[c]{MF \\ \cite{fu-sp16-a}} & \makecell[c]{NPE \\ \cite{wang-tip13-naturalness}} & \makecell[c]{SRIE \\ \cite{fu-cvpr16-a}} & \makecell[c]{SDD \\ \cite{hao-tmm20-low}} & \makecell[c]{CDEF \\\cite{lei-tmm22-low}} & \makecell[c]{LPNet \\\cite{li-tmm20-luminance}} \\
\hline
PSNR & 21.36 & 21.22 & 14.10 & 17.79 & 15.24 & 18.73 & 17.33 & 17.34 & 16.64 & 19.75 & 17.80\\
SSIM & 0.842 & 0.860 & 0.480 & 0.568 & 0.470 & 0.559 & 0.461 & 0.686 & 0.677 & 0.627 & 0.792\\
\hline
Methods & \makecell[c]{DRBN \\ \cite{yang-tip21-band}} & \makecell[c]{KinD \\ \cite{zhang-acmmm19-kindling}} & \makecell[c]{RUAS \\ \cite{liu-cvpr21-retinex}} & \makecell[c]{FIDE \\ \cite{xu-cvpr20-learning}} & \makecell[c]{EG \\ \cite{jiang-tip21-enlightengan}} & \makecell[c]{MS-RDN \\ \cite{yang-tip21-sparse}} & \makecell[c]{Retinex \\ -Net\cite{wei-bmvc18-deep}} & \makecell[c]{MIRNet \\ \cite{zamir-eccv20-learning}} & \makecell[c]{IPT \\\cite{chen-cvpr21-pre}} & \makecell[c]{Uformer \\\cite{wang-cvpr22-uformer}} & Ours \\
\hline
PSNR & 20.29 & 14.74 & 18.37 & 16.85 & 18.23 & 20.06 & 15.47 & 20.02 & 19.80 & 18.82 & \textbf{22.20}\\
SSIM & 0.831 & 0.641 & 0.723 & 0.678 & 0.617 & 0.816 & 0.567 & 0.820 & 0.813 & 0.771 & \textbf{0.863}\\
\hline
\hline
\end{tabular}
\end{center}
\caption{Quantitative comparisons with state-of-the-art methods on the LOL-v2-real dataset.}
\label{tab:quanti_comp_v2_real}
\end{table*}

\renewcommand{\tabcolsep}{5pt}
\begin{table*}[t]
\begin{center}
\begin{tabular}{cccccccccccc}
\hline
\hline
Methods & \makecell[c]{SNR \\ \cite{xu-cvpr22-snr}} & \makecell[c]{URetinex \\ -Net\cite{wu-cvpr22-uretinex}} & \makecell[c]{DeepLPF \\ \cite{moran-cvpr20-deeplpf}} & \makecell[c]{SCI \\ \cite{ma-cvpr22-toward}} & \makecell[c]{LIME \\ \cite{guo-tip16-lime}} & \makecell[c]{MF \\ \cite{fu-sp16-a}} & \makecell[c]{NPE \\ \cite{wang-tip13-naturalness}} & \makecell[c]{SRIE \\ \cite{fu-cvpr16-a}} & \makecell[c]{SDD \\ \cite{hao-tmm20-low}} & \makecell[c]{CDEF \\\cite{lei-tmm22-low}} & \makecell[c]{LPNet \\\cite{li-tmm20-luminance}} \\
\hline
PSNR & 23.84 & 18.75 & 16.02 & 17.73 & 16.88 & 17.50 & 16.59 & 14.50 & 16.46 & 18.56 & 19.51\\
SSIM & 0.905 & 0.829 & 0.587 & 0.763 & 0.776 & 0.751 & 0.778 & 0.616 & 0.728 & 0.842 & 0.846\\
\hline
Methods & \makecell[c]{DRBN \\ \cite{yang-tip21-band}} & \makecell[c]{KinD \\ \cite{zhang-acmmm19-kindling}} & \makecell[c]{RUAS \\ \cite{liu-cvpr21-retinex}} & \makecell[c]{FIDE \\ \cite{xu-cvpr20-learning}} & \makecell[c]{EG \\ \cite{jiang-tip21-enlightengan}} & \makecell[c]{MS-RDN \\ \cite{yang-tip21-sparse}} & \makecell[c]{Retinex \\ -Net\cite{wei-bmvc18-deep}} & \makecell[c]{MIRNet \\ \cite{zamir-eccv20-learning}} & \makecell[c]{IPT \\\cite{chen-cvpr21-pre}} & \makecell[c]{Uformer \\\cite{wang-cvpr22-uformer}} & Ours \\
\hline
PSNR & 23.22 & 13.29 & 16.55 & 15.20 & 16.57 & 22.05 & 17.13 & 21.94 & 18.30 & 19.66 & \textbf{25.58}\\
SSIM & 0.927 & 0.578 & 0.652 & 0.612 & 0.734 & 0.905 & 0.798 & 0.876 & 0.811 & 0.871 & \textbf{0.940}\\
\hline
\hline
\end{tabular}
\end{center}
\caption{Quantitative comparisons with state-of-the-art methods on the LOL-v2-synthetic dataset.}
\label{tab:quanti_comp_v2_synthetic}
\end{table*}

We visually show $\mathbf{R}_{s2}$ in Fig.~\ref{fig:exam_result}, observing that the illumination is enhanced and image details are also recovered. Especially, the noise is well removed by our COMO-ViT.

\vspace{-4mm}
\paragraph{LGCM.} LGCM takes $\mathbf{R}_{s2}$ as input, and learns local deep features to perceive illumination gap between $\mathbf{R}_{s2}$ and ground truth and elaborately enhances illumination to reduce local color deviation:
\begin{equation}\label{eq:lgcm_formu}
\begin{aligned}
\mathbf{\Gamma}_{l} = \phi(\operatorname{Conv}_3(\mathbf{R}_{s2})), \quad
\mathbf{R}_{s3} = \mathbf{R}_{s2}^{\mathbf{\Gamma}_{l}}.
\end{aligned}
\end{equation}
Similarly, we also apply Taylor Series to approximate exponential operation as follows:
\begin{equation}\label{eq:lgcm_taylor}
\begin{aligned}
\mathbf{R}_{s3} = \mathbbm{1} + \ln\mathbf{R}_{s2} \odot \mathbf{\Gamma}_{l} + \frac{\ln^{2}\mathbf{R}_{s2}}{2!} \odot \mathbf{\Gamma}^{2}_{l}.
\end{aligned}
\end{equation}
$\mathbf{\Gamma}_{l}$ and $\mathbf{R}_{3}$ are shown in Fig.~\ref{fig:exam_result}. We observe that the illumination is enhanced further, obtaining better visual quality.

\subsection{Loss Functions}
\label{sec:loss}
Our IAGC adopts a three-stage strategy to enhance low-light images. In each stage, we adopt the same loss functions to constrain the network training. The details are:
\begin{equation}\label{eq:loss}
\begin{aligned}
 \mathcal{L} = \sum^{3}_{i=1}( \sqrt{\Vert \mathbf{R}_{si} - \mathbf{G} \Vert^{2} + \epsilon^2} + \Vert\bigtriangledown\mathbf{R}_{si}- \bigtriangledown\mathbf{G}\Vert^2).
\end{aligned}
\end{equation}
The first term is the Charbonnier loss constraining network in the spatial domain. The second term is a gradient loss to supervise the network in the gradient domain, assisting the Charbonnier loss to recover more image details. $\mathbf{R}_{si}$ is the result of stage $i$, $\epsilon=10^{-3}$, $\mathbf{G}$ is ground truth, $\bigtriangledown$ denotes the combination of the horizontal and vertical gradients.

%-------------------------------------------------------------------------
\section{Experiments}
\label{sec:expes}

\subsection{Datasets and Experimental Setting}

\vspace{-0mm}
\paragraph{Datasets.} We evaluate our method on the widely-used LOL (including v1 and v2) datasets. LOL-v1 \cite{wei-bmvc18-deep} contains $485$ real low-/normal-light pairs for training and $15$ pairs for testing. LOL-v2 \cite{yang-tip21-sparse} is composed of two sets. LOL-v2-real contains $689$ pairs for training and $100$ pairs for testing. The majority of the low-light images are collected by changing exposure time and ISO while fixing other camera configurations. LOL-v2-synthetic is collected by analyzing the illumination distribution so that synthetic images match the property of real dark photography. This set contains $900$ synthetic pairs for training and $100$ pairs for testing.

\vspace{-4mm}
\paragraph{Experimental Setting.} Our IAGC adopts an end-to-end training strategy to optimize the network parameters of $3$ stages simultaneously. Our model is implemented with PyTorch \cite{paszke-nips19-pytorch} and trained on an NVIDIA v100NV32 GPU. Other hyper-parameters of training and network structure are listed in Table \ref{tab:hyper_para} for clarity.

\subsection{Comparison with Current Methods}
We compare the results of the proposed IAGC with a rich collection of SOTA methods to make comprehensive quantitative comparisons. For qualitative comparison, we select some very recent or robust methods to visually show their performance, including SNR \cite{xu-cvpr22-snr}, URetinex-Net \cite{wu-cvpr22-uretinex}, DeepLPF \cite{moran-cvpr20-deeplpf}, SCI \cite{ma-cvpr22-toward}, LIME \cite{guo-tip16-lime}, MF \cite{fu-sp16-a}, NPE \cite{wang-tip13-naturalness}, SRIE \cite{fu-cvpr16-a}, SDD \cite{hao-tmm20-low}, CDEF \cite{lei-tmm22-low}, LPNet \cite{li-tmm20-luminance}, DRBN \cite{yang-tip21-band}, KinD \cite{zhang-acmmm19-kindling}, RUAS \cite{liu-cvpr21-retinex}, FIDE \cite{xu-cvpr20-learning}, EG \cite{jiang-tip21-enlightengan}, MS-RDN \cite{yang-tip21-sparse}, Retinex-Net \cite{wei-bmvc18-deep}, MIRNet \cite{zamir-eccv20-learning}. Moreover, following \cite{xu-cvpr22-snr}, another two classical architecture IPT \cite{chen-cvpr21-pre} and Uformer \cite{wang-cvpr22-uformer} are also selected to make more extensive comparisons.

\vspace{-4mm}
\paragraph{Quantitative Comparisons.} We use PSNR and SSIM \cite{wang-tip04-image} as quantitative metrics. Table~\ref{tab:quanti_comp_v1}, \ref{tab:quanti_comp_v2_real} and \ref{tab:quanti_comp_v2_synthetic} show that our method obtains consistent best values. Especially on the LOL-v2-real and LOL-v2-synthetic datasets, our method surpasses the second best SNR \cite{xu-cvpr22-snr} by large margins. Note that these numbers are obtained either by running their respective codes or from their respective papers.

\vspace{-4mm}
\paragraph{Visual Quality Comparisons.} Due to space limitation, we select the very recent URetinexNet \cite{wu-cvpr22-uretinex}, SCI \cite{ma-cvpr22-toward}, SNR \cite{xu-cvpr22-snr} and two classic LIME \cite{guo-tip16-lime} and NPE \cite{wang-tip13-naturalness} to visually make comparison with our method. Fig.~\ref{fig:real_comp_rain} shows three low-light images respectively from LOL-v1 (the first line), LOL-v2-real (the second line), and LOL-v2-synthetic (the third line). We observe that LIME and NPE recover good image details and color hue for the first image, but apparent noise remains. They do not produce good results on the second and third images. SCI may produce slight noise, e.g., in the first and second images. Moreover, this method does not recover the illumination well. URetinexNet obtains good performance on the first and second images but does not perform well on the third synthetic image. SNR performs well on all these three images. Especially, this method handles the noise well. By comparison, our method enhances better illumination and preserves more details.

\renewcommand{\tabcolsep}{3pt}
\begin{table}[t]
\begin{center}
\begin{tabular}{lcccccc}
\hline
\hline
\multirow{2}{*}{Datasets} & \multicolumn{2}{c}{$\mathcal{G}_{1}$} & \multicolumn{2}{c}{$\mathcal{G}_{2}$} & \multicolumn{2}{c}{$\mathcal{G}_{3}$} \\
\cmidrule(r){2-3} \cmidrule(r){4-5} \cmidrule(r){6-7} 
& PSNR & SSIM & PSNR & SSIM & PSNR & SSIM \\
\hline
LOL-v1 & 23.18 & 0.821 & 23.96 & 0.838 & 24.53 & 0.842  \\
LOL-v2-real & 21.32 & 0.834 & 21.87 & 0.851 & 22.20 & 0.863 \\
LOL-v2-syn & 24.88 & 0.912 & 25.26 & 0.925 & 25.58 & 0.940 \\
\hline
\hline
\end{tabular}
\end{center}
\caption{Quantitative results of our ablation study
on the proposed gamma correction modules GGCM and LGCM.}
\label{tab:ablate_gamma}
\end{table}

\begin{figure}[t]
\begin{center}
\begin{minipage}{0.19\linewidth}
\centering{\includegraphics[width=1\linewidth]{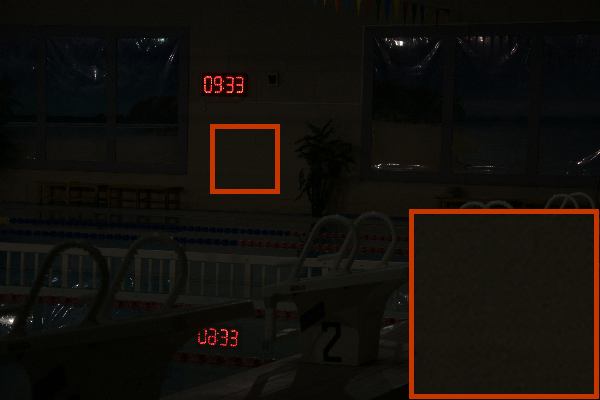}}
\centerline{Input}
\end{minipage}
\hfill
\begin{minipage}{0.19\linewidth}
\centering{\includegraphics[width=1\linewidth]{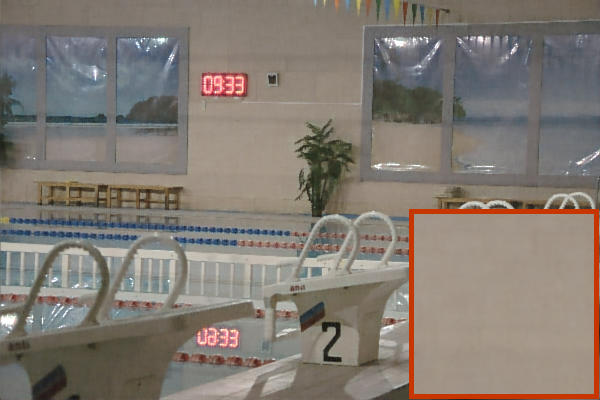}}
\centerline{$\mathcal{G}_{1}$}
\end{minipage}
\hfill
\begin{minipage}{0.19\linewidth}
\centering{\includegraphics[width=1\linewidth]{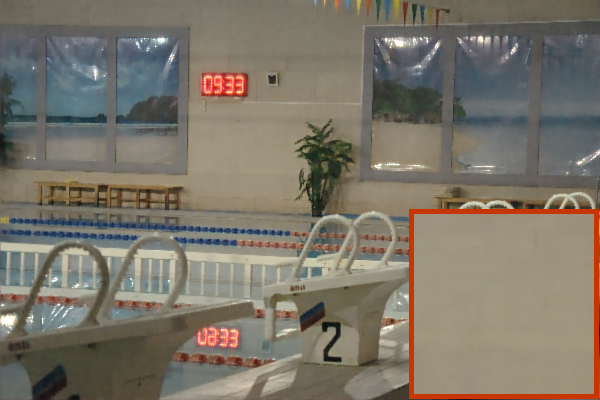}}
\centerline{$\mathcal{G}_{2}$}
\end{minipage}
\hfill
\begin{minipage}{0.19\linewidth}
\centering{\includegraphics[width=1\linewidth]{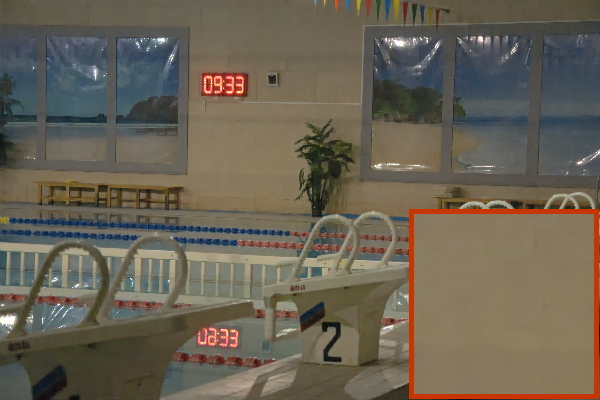}}
\centerline{$\mathcal{G}_{3}$}
\end{minipage}
\hfill
\begin{minipage}{0.19\linewidth}
\centering{\includegraphics[width=1\linewidth]{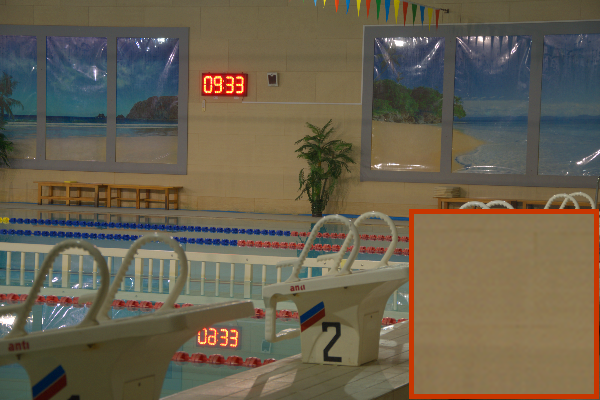}}
\centerline{GT}
\end{minipage}
\end{center}
\caption{Qualitative comparisons of our ablation study on the proposed GGCM and LGCM.}
%\vspace{-6mm}
\label{fig:ablate_gamma}
\end{figure}

\subsection{Ablation Studies}

\vspace{-0mm}
\paragraph{Effect of Gamma Correction Modules.} Illumination-aware gamma correction modules, i.e., GGCM and LGCM play important roles in illumination enhancement. GGCM is to perceive the whole illumination and adaptively predict a global gamma correction factor to enhance brightness. LGCM is aware of illumination differences by exploring local pixel dependencies and predicts gamma correction factors pixel-wisely. We verify their effect by ablating them separately. $\mathcal{G}_{1}$ is the model ablating GGCM, $\mathcal{G}_{2}$ is the one removing LGCM, $\mathcal{G}_{3}$ is our whole network. As shown in Table~\ref{tab:ablate_gamma}, SSIM does not have too large differences for these three cases. The PSNR of $\mathcal{G}_{1}$ decreases more largely than that of $\mathcal{G}_{2}$, illustrating that GGCM plays a more important role in illumination enhancement than LGCM. Fig.~\ref{fig:ablate_gamma} visually shows similar results, when ablating GGCM, the hue deviates more apparently from that of ground truth.

\renewcommand{\tabcolsep}{7pt}
\begin{table}[t]
\begin{center}
\begin{tabular}{lcccc}
\hline
\hline
Datasets & $\mathcal{A}_{1}$ & $\mathcal{A}_{2}$ & $\mathcal{A}_{3}$ & $\mathcal{A}_{4}$ \\
\hline
LOL-v1 & \makecell[c]{23.72\\0.797} & \makecell[c]{24.13\\0.816} & \makecell[c]{24.21 \\ 0.829} & \makecell[c]{24.53\\0.842}  \\
\hline
LOL-v2-real & \makecell[c]{21.54\\0.811} & \makecell[c]{21.93\\0.834} & \makecell[c]{22.04 \\ 0.851} & \makecell[c]{22.20\\0.863} \\
\hline
LOL-v2-syn & \makecell[c]{24.89\\0.902} & \makecell[c]{25.27\\0.917} & \makecell[c]{25.31\\0.923} & \makecell[c]{25.58 \\ 0.940} \\
\hline
\hline
\end{tabular}
\end{center}
\caption{Quantitative results (PSNR/SSIM) of our ablation study on the proposed local-to-global self-attention mechanism.}
\label{tab:ablate_attention}
\end{table}

\begin{figure}[t]
\begin{center}
\begin{minipage}{0.32\linewidth}
\centering{\includegraphics[width=1\linewidth]{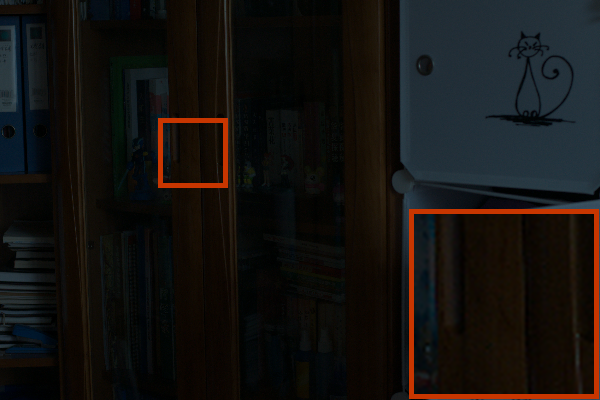}}
\centerline{Input}
\end{minipage}
\begin{minipage}{0.32\linewidth}
\centering{\includegraphics[width=1\linewidth]{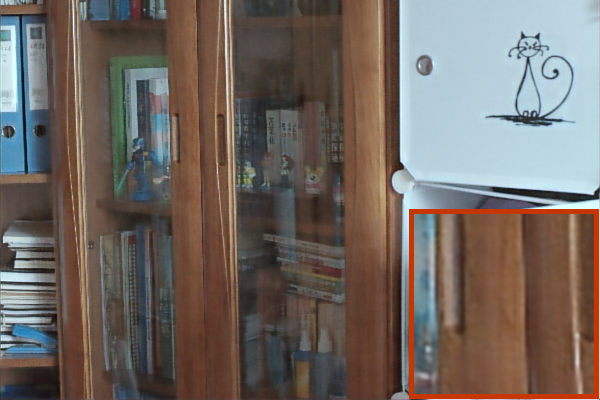}}
\centerline{$\mathcal{A}_{1}$}
\end{minipage}
\begin{minipage}{0.32\linewidth}
\centering{\includegraphics[width=1\linewidth]{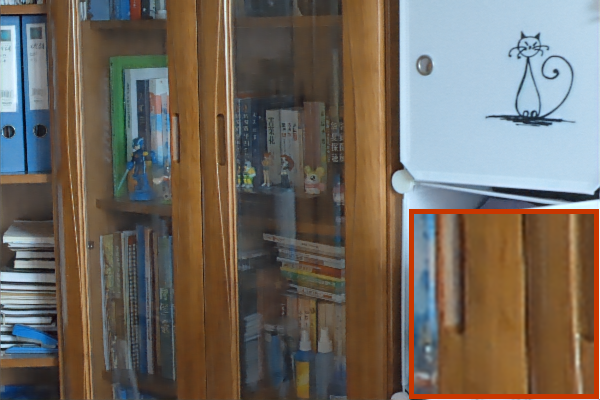}}
\centerline{$\mathcal{A}_{2}$}
\end{minipage}
\vfill
\begin{minipage}{0.32\linewidth}
\centering{\includegraphics[width=1\linewidth]{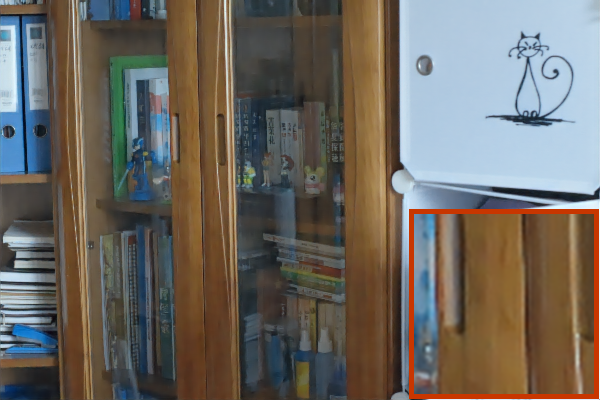}}
\centerline{$\mathcal{A}_{3}$}
\end{minipage}
\begin{minipage}{0.32\linewidth}
\centering{\includegraphics[width=1\linewidth]{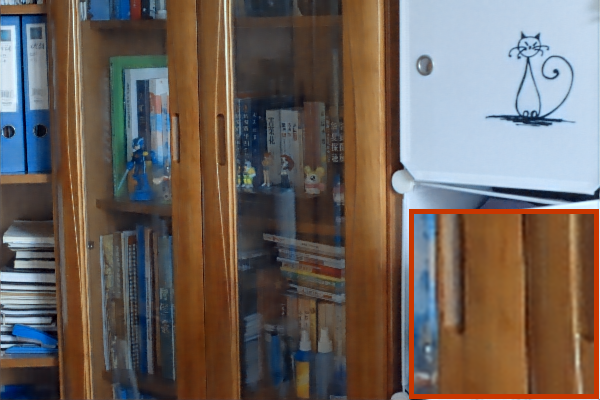}}
\centerline{$\mathcal{A}_{4}$}
\end{minipage}
\begin{minipage}{0.32\linewidth}
\centering{\includegraphics[width=1\linewidth]{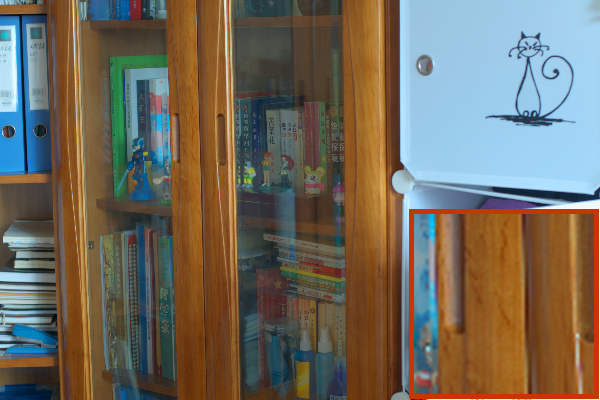}}
\centerline{GT}
\end{minipage}
\end{center}
\caption{Qualitative comparisons of our ablation study on the proposed local-to-global self-attention mechanism.}
\label{fig:ablate_atten}
\vspace{-4mm}
\end{figure}

\vspace{-4mm}
\paragraph{Effect of Local-to-Global Self-Attention.} COMO-ViT is designed to completely model the dependencies of all pixels via a local-to-global hierarchical self-attention mechanism. The local attention is to explore pixel relationships in a window. The global attention connects different windows so that all pixels across the whole image could be combined together and completely modeled. Here, we test their effect on illumination enhancement. We define $\mathcal{A}_{1}$ as the model ablating local attention, which degrades to a patch-embedding Transformer. $\mathcal{A}_{2}$ is the model removing global self-attention, which acts like SwinIR \cite{liang-iccvw21-swinir}. $\mathcal{A}_{3}$ combines local and global attention as in TNT \cite{han-nips21-transformer}. $\mathcal{A}_{4}$ is our whole network. Table~\ref{tab:ablate_attention} shows that performances of $\mathcal{A}_{1}$ and $\mathcal{A}_{2}$ both decrease. But $\mathcal{A}_{1}$ possesses lower performance, illustrating that local self-attention plays a more important role. The performance of $\mathcal{A}_{4}$ is higher than that of $\mathcal{A}_{3}$, which verifies that our local-to-global hierarchical self-attention mechanism is more effective than the bypass architecture in TNT. The reason is that our COMO-ViT better expresses images via pixel-level self-attention, the dark areas can be better inferred by using far informative bright areas. The visual results are in Fig.~\ref{fig:ablate_atten}.

\renewcommand{\tabcolsep}{3pt}
\begin{table}[t]
\begin{center}
\begin{tabular}{lcccccc}
\hline
\hline
\multirow{2}{*}{Win size ($w$)} & \multicolumn{2}{c}{$4$} & \multicolumn{2}{c}{$16$} & \multicolumn{2}{c}{$32$} \\
\cmidrule(r){2-3} \cmidrule(r){4-5} \cmidrule(r){6-7} 
& PSNR & SSIM & PSNR & SSIM & PSNR & SSIM \\
\hline
LOL-v1 & 24.46 & 0.836 & 24.53 & 0.842 & 24.37 & 0.830  \\
LOL-v2-real & 22.14 & 0.857 & 22.20 
 & 0.863 & 22.23 & 0.851 \\
LOL-v2-syn & 25.44 & 0.931 & 25.58 & 0.940 & 25.39 & 0.935 \\
\hline
\hline
\end{tabular}
\end{center}
\caption{Quantitative results of different window sizes.}
\label{tab:ablate_winsize}
\end{table}

\begin{figure}[t]
\begin{center}
\begin{minipage}{0.19\linewidth}
\centering{\includegraphics[width=1\linewidth]{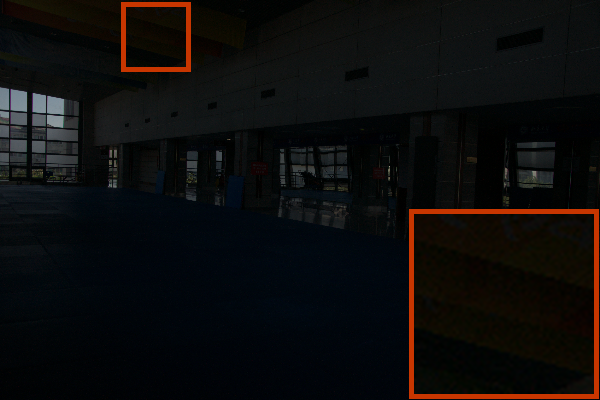}}
\centerline{Input}
\end{minipage}
\begin{minipage}{0.19\linewidth}
\centering{\includegraphics[width=1\linewidth]{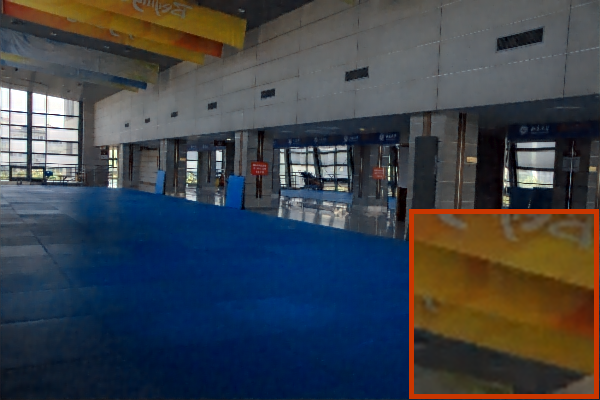}}
\centerline{$w=4$}
\end{minipage}
\begin{minipage}{0.19\linewidth}
\centering{\includegraphics[width=1\linewidth]{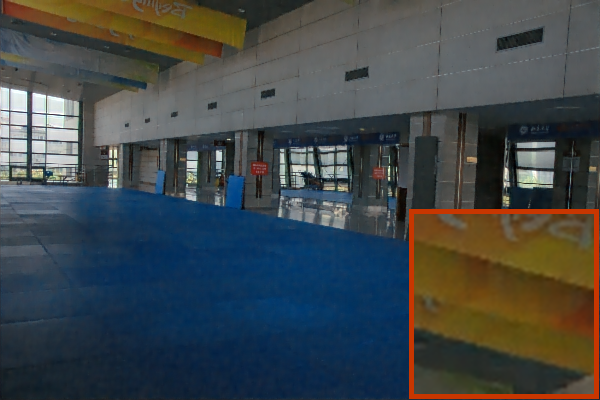}}
\centerline{$w=16$}
\end{minipage}
\begin{minipage}{0.19\linewidth}
\centering{\includegraphics[width=1\linewidth]{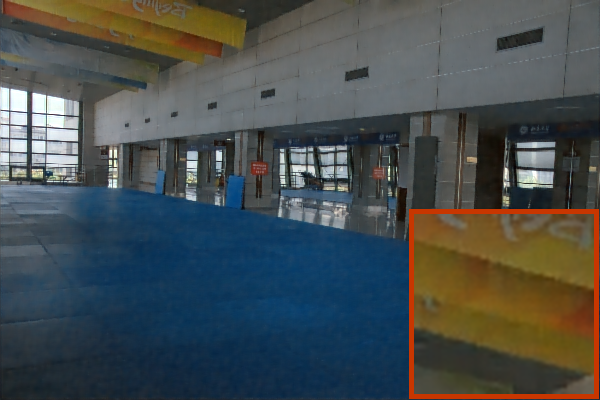}}
\centerline{$w=32$}
\end{minipage}
\begin{minipage}{0.19\linewidth}
\centering{\includegraphics[width=1\linewidth]{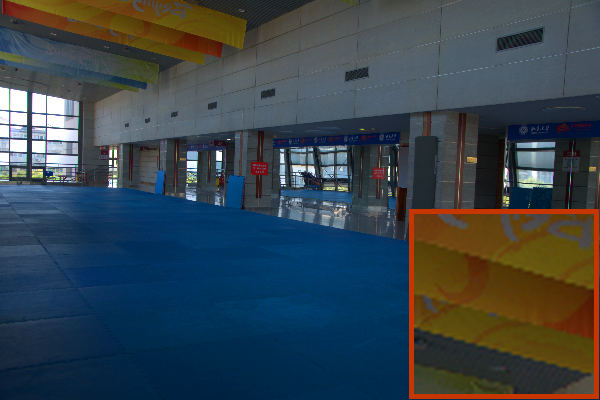}}
\centerline{GT}
\end{minipage}
\end{center}
\caption{Visual comparison results for different window sizes.}
\vspace{-4mm}
\label{fig:ablate_winsize}
\end{figure}

\vspace{-4mm}
\paragraph{Effect of Window Size.} In COMO-ViT, the input image is split into $w \times w$ windows, the default $w$ is $16$. Because the value of $w$ also influences the embedding dimension of our global self-attention, here we test the effect of other values of $w$ on the model performance. Table~\ref{tab:ablate_winsize} and Fig.~\ref{fig:ablate_winsize} both show that $w$ just has slight influence on the performance, we thus set $w=16$ for its efficiency ($0.735M$ parameters for $w=4$; $0.326M$ for $w=16$; $1.016M$ for $w=32$).

\vspace{-4mm}
\paragraph{Limitations.} Although IAGC achieves better performance than previous methods, there are still some open problems that IAGC does not solve. There exists slight local color deviation when encountering very challenging low-light cases, e.g., the color gaps at the upper right corner in Fig.~\ref{fig:ablate_atten}. This can be understood as contrast and hue are seriously damaged and accurate contextual relationships are not available to recover complete image color. This is a limitation of all existing methods, including IAGC. 

\section{Conclusion}
In this paper, we propose a novel Illumination-Aware Gamma Correction (IAGC) network for low-light image enhancement. In particular, we combine the effectiveness of gamma correction with the strong modelling capacities of networks and introduce two modules to predict the global and local gamma correction factors via perceiving the global (GGCM) and local (LGCM) illumination. Moreover, in order to solve the large-scale dark areas in low-light images, we design a novel Transformer block (COMO-ViT) to completely model the dependencies of all pixels across images through a local-to-global self-attention mechanism. The local self-attention explores pixel-level dependencies based on local windows, while the global self-attention further expands pixel dependencies to the whole image. Compared with conventional Transformers which either only focus on local information but discard global dependencies or emphasize global relationships but neglect local information, our IAGC recovers better illumination and preserves more image details. Extensive experiments have demonstrated the effectiveness of the proposed IAGC architecture. 

\noindent\textbf{Acknowledgements} This work was supported by Sichuan Science and Technology Program of China under grant No.2023NSFSC0462. 

{\small
\bibliographystyle{ieee_fullname}
\bibliography{final_version}
}

\end{document}